\pdfoutput=1

\documentclass[11pt]{article}

\usepackage[preprint]{acl}

\usepackage{times}
\usepackage{latexsym}

\usepackage[T1]{fontenc}

\usepackage[utf8]{inputenc}

\usepackage{microtype}

\usepackage{inconsolata}

\usepackage{graphicx}

\usepackage{algorithm}
\usepackage{algorithmic}
\usepackage{xcolor}
\usepackage{xspace}
\usepackage{enumitem}
\usepackage{rotating, multirow}
\usepackage{subcaption}     
\usepackage{bold-extra}     
\usepackage{bm}             
\usepackage[hang,flushmargin]{footmisc}  
\usepackage{stfloats}  
\usepackage{amsfonts,amsmath,amssymb}
\usepackage{pifont}
\usepackage{array,booktabs,makecell,tabularx}
\usepackage{lipsum}
\usepackage{csquotes}
\usepackage{changepage}
\usepackage{CJKutf8}
\newcommand\LN{\linebreak\noindent}
\newcommand{\cjk}[1]{\begin{CJK*}{UTF8}{gkai}#1\end{CJK*}}

\title{Dense Retrievers Can Fail on Simple Queries:\\Revealing The Granularity Dilemma of Embeddings}

\author{
Liyan Xu$^1$~~~~Zhenlin Su$^2$~~~~Mo Yu$^1$~~~~Jiangnan Li$^1$\\ \bf Fandong Meng$^1$~~~~~Jie Zhou$^1$ \\
$^1$Pattern Recognition Center, WeChat AI\\
$^2$South China University of Technology\\
\small{\texttt{liyanlxu@tencent.com}}}

\begin{document}
\maketitle

\begin{abstract}
This work stems from an observed limitation of text encoders: embeddings may not be able to recognize fine-grained entities or events within encoded semantics, resulting in failed retrieval even in simple cases. To examine such behaviors, we first introduce a new evaluation dataset, CapRetrieval, in which passages are image captions and queries are phrases targeting entity or event concepts in diverse forms. Zero-shot evaluation suggests that encoders often struggle with these fine-grained matching, regardless of training sources or model size. Aiming for enhancement, we proceed to finetune encoders with our proposed data generation strategies, enabling a small 0.1B encoder to outperform the state-of-the-art 7B model. Within this process, we further uncover the \emph{granularity dilemma}, a challenge for embeddings to capture fine-grained salience while aligning with overall semantics.
Our dataset, code and models in this work are publicly released at \url{https://github.com/lxucs/CapRetrieval}.
\end{abstract}
\section{Introduction}

Dense Passage Retrieval (DPR) is a crucial component in searching, characterized by the classic dual-encoder paradigm \cite{sbert}: queries and candidate passages are independently encoded to embeddings that capture their semantics, then top candidates are retrieved based on their embedding similarity.
For large language models (LLMs) especially, DPR serves a pivotal role in Retrieval Augmented Generation (RAG) \cite{rag}. The training of such encoders has thereby become an active direction in representation learning \cite{repr-survey}.

However, despite their growing capabilities in encoding complex queries and documents, embedding matching can still fail on simple cases, and yet unable to fully supersede the conventional lexical-based methods such as BM25 \cite{10.1145/3459637.3482159,ren-etal-2023-thorough}. Consequently, retrieval systems in practice usually take a hybrid approach for optimized performance \cite{spar,hybrid-retriever,light-hybrid,m3}. 
In this work, we probe into such limitation of DPR encoders\footnote{We focus on embedding-based DPR in this work, and leave reranking or other hybrid paradigms as a separate scope.}, and highlight that even for simple passages, state-of-the-art encoders still often \textbf{lack a fine-grained view of concepts} and the integration of world knowledge, resulting in failed retrieval on entities or events.
To remedy this issue, we delve into data generation strategies towards enhanced encoder training, and we identify a \emph{granularity dilemma} within this process, posing a challenge for general training data composition.

Illustrated by Table~\ref{tab:example-a}\&\ref{tab:example-b}, with strong encoders of both BGE \cite{bge} and GTE \cite{mgte} series, spanning from 0.1B to 7B models that rank top on the MTEB leaderboard\footnote{\url{https://huggingface.co/spaces/mteb/leaderboard}} for Chinese, all their embeddings fail to rank the more relevant passages against obviously less relevant ones, for the simple query \emph{fried chicken}, \emph{purple flower}, and \emph{watermelon} respectively, indicating that such phenomena occur universally, regardless of training sources, model sizes or languages.

To facilitate the analysis of such behaviors and potential improvements, we first construct a new evaluation dataset in both Chinese and English, tailored to a practical image search scenario, where the candidate passages are image captions, and queries are short phrases of entities or events reflected in captions.
The dataset overall comprises seemingly straightforward queries and captions, dubbed \textbf{CapRetrieval}, and features two unique aspects. First, it naturally requires fine-grained semantic encoding of passages, inquiring concepts of entities or events reflected in the captions. Second, distinguished from prior caption-related datasets, e.g. Ecom/Video Retrieval \cite{multicpr}, whose labels are collected from real-world user clicks and thus may contain many false negatives, CapRetrieval provides explicit annotations for each query-passage pair, enabling more reliable and in-depth analysis by near 1.3 million pair labels.

Upon our zero-shot evaluation on CapRetrieval using open-source encoders of various sizes, all of them are revealed certain flaws performing these fine-grained matching, underscoring the current limitations of embeddings. 
We then proceed to finetune the encoder with our proposed data generation strategies in Sec.~\ref{sec:train}, where we leverage LLMs for automatic keyword generation as training data to enforce the semantic encoding of precise salience.

Empirical results show that our finetuned 0.1B encoder outperforms all baselines on CapRetrieval, surpassing 7B encoders, validating that the data generation strategies are effective strengthening a more subtle semantic matching.
Nevertheless, we also identify a dilemma upon further analysis: with the introduced keywords added into training, the model improves on granular saliency and concepts, but may lose grasp on the overall critical semantics in a bigger picture, resulting in degradation on certain scenarios. We describe it as the \emph{granularity dilemma} to handle both coarse and fine granularity (Sec.~\ref{sec:train}), showing that it still remains a challenge to obtain general embeddings of a full semantic view.

\section{Dataset: CapRetrieval}
\label{sec:dataset}

Our dataset CapRetrieval is constructed for typical retrieval evaluation towards a practical image search scenario, composed of three following parts.

\paragraph{Passages}
A pool of image captions are prepared as candidate passages, derived from three steps. (1) A set of images is collected from personal phone albums for diverse coverage of different types, including daily photographs, web pictures, screenshots of various apps or articles. (2) Each image is transcribed into a short caption by prompting GPT-4o. (3) Each caption is manually reviewed and anonymized to ensure privacy compliance.

\paragraph{Queries}
User queries are short phrases to search relevant images, collected by two rounds. The first round of queries are brainstormed by researchers of this work, being usual entities or events that align with real-world usage scenarios. For the second round, a combination of queries in the first round is performed, then manually revised to form more complex queries. Among these queries, the easiest ones can be resolved by lexical matching, while others require fine-grained concept understanding and world knowledge to resolve.

\paragraph{Labels}
Following the convention of typical retrieval datasets in MTEB, the label is a relevance score of $[0,1,2]$ for a pair of query and passage, signaling no/weak/strong relevance. Unlike prior web-collected datasets with large-scale passage pools, our controlled setting enables exhaustive annotations for each query-passage pair, which minimizes false negatives for a more reliable evaluation.

Note that the annotation and the entire retrieval process only concern textual captions, without involving vision features. The accuracy of image captioning thus does not affect the evaluation.
\begin{itemize}[noitemsep,nolistsep]
    \item Strong Relevance ($2$): the query is certain or almost certain directly reflected in the caption according to commonsense. For example, ``\emph{the evening sky filled with dense clouds, with sunlight streaming through them}'' is labeled $2$ to the query ``\emph{sunset}''.
    \item Weak Relevance ($1$): the query is likely but not certainly reflected in the caption, or the query is not directly reflected but indeed highly related. For example, ``\emph{a floor plan of a 2B2B apartment of approximately 90.89 square meters}'' does not directly mention the action ``\emph{real estate purchase}'', but is labeled $1$ due to high relatedness by search intents.
    \item No Relevance ($0$): those who do not meet up with the strong or weak relevance.
\end{itemize}

\begin{table}[tbp!]
\centering
\resizebox{0.82\columnwidth}{!}{
\begin{tabular}{l|c|ccc}
\toprule
& Records & Min & Max & Avg \\
\midrule
Passages & 3024 & 8 & 75 & 30.90\\
Queries & 404 & 1 & 16 & 3.78 \\
\bottomrule
\end{tabular}}
\caption{Basic statistics of CapRetrieval dataset: number of records, and min/max/avg number of tokens per passage and query. The full annotations of all 1.3M query-passage pairs comprise 4,683 positive pairs.}
\label{tab:datasets}
\vspace{-1.5ex}
\end{table}

\paragraph{Statistics}
Table~\ref{tab:datasets} depicts the basic statistics of CapRetrieval. 
For analysis, we roughly categorize all queries in CapRetrieval into eight types, of which four types (\emph{object}, \emph{person}, \emph{place}, \emph{concept}) together can be regarded as \emph{Singleton Entity}. Both \emph{Singleton Entity} and \emph{Singleton Event} are relatively straightforward phrases inquiring entities/events without imposing extra conditions, whereas other query types involve more complex constraints and semantics.
More details and statistics on query types and labels are described in Appx.~\ref{app:dataset}.

CapRetrieval was originally constructed in Chinese. We further employ GPT-4o to translate all queries and passages into English to provide the corresponding English version, of which the translation quality is verified by human experts. The remaining sections in this paper reports on the original CapRetrieval without further specification.

\section{Zero-Shot Evaluation}
\label{sec:zero-shot}

To examine the performance of off-the-shelf encoders on CapRetrieval, we evaluate five popular open-source encoder series for Chinese on Huggingface, offering various model sizes:

• BGE\footnote{bge-\{base,large\}-zh-v1.5} \cite{bge}: 0.1B / 0.3B

• GTE\footnote{gte-multilingual-base;\; gte-Qwen2-\{1.5B,7B\}-instruct} \cite{mgte}: 0.1B / 1.5B / 7B

• E5\footnote{multilingual-e5-\{base,large\};\; e5-mistral-7B-instruct} \cite{e5,e5-mistral}: 0.1B / 0.3B / 7B

• Conan-v1\footnote{Conan-embedding-v1 (v2 has not been released yet)} \cite{conan-v1}: 0.3B

• Qwen3\footnote{Qwen3-Embedding-\{0.6B,8B\}} \cite{zhang2025qwen3embeddingadvancingtext}: 0.6B / 8B

\noindent Our experimental settings comply with the retrieval protocol in MTEB \cite{mteb}, adopting nDCG@10 as the main evaluation metric. Additionally, since full labels of all pairs are annotated for CapRetrieval, we also provide nDCG@1/5 for more precise evaluation. For each model, we follow the recommended usage by its publishers' instructions. Performance of BM25 is also provided as a baseline for reference. More implementation details are described in Appx.~\ref{app:zero-shot}.

\begin{table}[tbp!]
\centering
\resizebox{\columnwidth}{!}{
\begin{tabular}{ll|ccc}
\toprule
&& nDCG@1 & nDCG@5 & \bf nDCG@10 \\
\midrule
& \it BM25 & 74.40 & 69.30 & 66.54 \\
\midrule
\multirow{3}{*}{0.1B} & BGE & 81.30 & 78.97 & 78.86 \\
& GTE & 82.49 & 80.48 & 79.67 \\
& E5 & 80.11 & 77.31 & 76.33 \\
\midrule
\multirow{3}{*}{0.3B} & BGE & 83.42 & 78.94 & 79.15 \\
& E5 & 82.76 & 81.17 & 81.01 \\
& Conan-v1 & 78.78 & 77.30 & 77.04 \\
\midrule
\multirow{1}{*}{0.6B} & Qwen3 & 85.41 & 81.14 & 81.04 \\
\midrule
\multirow{4}{*}{> 1B} & GTE-1.5B & 81.70 & 77.20 & 77.35 \\
& GTE-7B & \bf 89.12 & \bf 86.94 & \bf 86.55 \\
& E5-7B & 77.59 & 76.02 & 76.40 \\
& Qwen3-8B & 87.00 & 84.95 & 84.61 \\
\midrule
& \it Human & 100.00 & 98.57 & 97.83 \\
\bottomrule
\end{tabular}}
\caption{Evaluation results of zero-shot experiments on CapRetrieval, with encoders of different model sizes. Human performance is evaluated on a 10\% subset. Evaluation on the according English dataset (CapRetrievalEn) is separately presented in Table~\ref{tab:zero-result-en}.}
\label{tab:zero-result}
\end{table}

\paragraph{Results}
Table~\ref{tab:zero-result} shows the results of zero-shot evaluation. As most queries and captions in CapRetrieval do not possess complex semantics, all models are able to achieve decent scores as expected, with nDCG@10 above 76. The best performance is obtained by GTE-7B with 86.55 nDCG score. Several observations can be further made:

• \textbf{All encoders exhibit flaws} matching these fine-grained queries, even the capable 7B/8B LLM models. Though, they all outperform BM25 by at least 10\%, underscoring that lexical matching alone is not able to resolve this task.

• \textbf{Model size is not the principal factor}. Within the same GTE series, the much smaller 0.1B model even outperforms the 1.5B model, and falls behind the 7B by 7\%, despite the huge size difference.

\paragraph{Query Analysis}
Table~\ref{tab:analysis} shows the decomposed performance on CapRetrieval by query types (described in \ref{app:query-type}). \emph{Singleton Entity} has the highest nDCG score as 82.05, while \emph{Singleton Event} and \emph{Simple Condition} have low scores as 73.2 and 73.8 respectively. The overall trend suggests that the encoder performs worse on more abstract queries, i.e. events or phrases with conditions. Entity-centric queries are relatively easier to resolve (80+ nDCG), though not by a large margin.

\paragraph{Embedding vs. BM25}
The right section of Table~\ref{tab:analysis} presents a comparison between embedding and BM25. BM25 exhibits more polarized performance, where it outperforms embedding on entity-centric queries, but lags behind on more abstract queries. The limitation of BM25 is especially pronounced for \emph{Complex Condition}, with a 66\% gap, highlighting the \textbf{necessity of embedding-based retrieval}. Overall, it also reveals room for improvement in embeddings as follows.

\begin{table}[tbp!]
\centering
\resizebox{0.96\columnwidth}{!}{
\begin{tabular}{l|c|ccc}
\toprule
& nDCG@10 & E>B & E<B & E=B \\
\midrule
\it Singleton Entity & \bf 82.05 & 28\% & \bf 40\% & \bf 32\% \\
\it Singleton Event & 73.21 & 50\% & 25\% & 25\% \\
\it Conjunction & 80.60 & 38\% & 38\% & 25\% \\
\it Simple Cond. & 73.80 & 58\% & 20\% & 22\% \\
\it Complex Cond. & 77.30 & \bf 73\% & 7\% & 20\% \\
\bottomrule
\end{tabular}}
\caption{Zero-shot performance of BGE 0.1B encoder per query type. The right part depicts the comparison with BM25: the ratio of queries when \textbf{E}mbeddings obtain higher/lower/similar (>/</=) scores than/to \textbf{B}M25.}
\label{tab:analysis}
\vspace{-1.5ex}
\end{table}

\paragraph{False Negatives}
We identify three error types as the common shortcomings of current embeddings.

• \emph{Direct miss}: embeddings may miss  entities or events reflected directly in passages, indicating that the current embedding \textbf{lacks a full view of semantics}, despite queries and passages are relatively short already. Errors can be further divided into two types within this scope.

i) \emph{Literal Error}: embeddings fail to retrieve passages that contain the full or partial query terms verbatim - passages that BM25 can successfully recall. Though these cases can be remedied by adding lexical search in practice, we advocate that embeddings should encode concepts with full information view. Resolving these seemingly simple matches is still challenging for state-of-the-art encoders, which can be regarded as the \emph{embedding} version of the ``needle in a haystack'' test for LLMs.

ii) \emph{Semantic Error}: the query is reflected by paraphrasing or in a more abstract way in passages, where embeddings generally outperform BM25 but still have room for improvement.

• \emph{Taxonomy knowledge}: certain scenarios require taxonomy knowledge involved, e.g. when inquiring a hypernym term such as ``\emph{household appliances}'' or ``\emph{seafood}'', or to recognize the matching between ``\emph{Cantonese-style roasted meats}'' and ``\emph{char siu}''.

• \emph{Commonsense reasoning}: some matches require commonsense reasoning to resolve, for instance, the encoder needs to know the color of \emph{lavender} to correctly handle the query ``\emph{purple flower}'', or to realize the mention of \emph{sitting} in a passage is highly relevant to the query ``\emph{chair}''.

\paragraph{False Positives}
We further summarize two error types that appear common in false positives.

• \emph{Over-generalization}: the passage contains relevant elements or shared tokens as the query, but does not reflect the actual query itself. For instance, the query ``\emph{shoe}'' retrieves ``\emph{a labeled cardboard box with an anti-trample symbol}'' before more relevant captions that actually mention shoes; ``\emph{subway}'' ranks captions regarding ``\emph{high-speed train ticket}'' before the ones mentioning subway stations.

• \emph{Ignoring subjects or conditions}: the passage only addresses partial semantics but fails to accommodate full concepts, e.g. ``\emph{purple flower}'' retrieves captions about \emph{purple butterflies}; ``\emph{shopping cart screenshot}'' retrieves a screenshot but of a ride-hailing app. As this kind of errors are quite common for queries with conditions, it suggests that \textbf{embeddings may not actually encode concepts}, but in a way towards superficial matching.

\section{Encoder Training}
\label{sec:train}

The results of zero-shot evaluation on CapRetrieval calls for more expressive embeddings that capture fine-grained concepts and world knowledge integration.
We proceed with further examination by finetuning encoders with training data strategies.

\subsection{Training Data Generation}
\label{ssec:train-data}

Training pairs in existing large-scale resources such as mMARCO \cite{mmarco} and DuReader \cite{dureader-rt} mainly comprise user search queries and clicks. Consequently, for a passage, the queries associated with a given passage in the training set are often coarse-grained, such that they do not address the full semantic content. Motivated towards fine-grained semantic matching, we propose automatic query generation for enhanced training, with distinct granularity as follows.

\begin{itemize}[noitemsep,nolistsep,leftmargin=*]
\item Overall \underline{s}um\underline{m}aries (\textbf{\texttt{SM}}): we ask LLMs to generate summaries and long questions regarding a passage, focusing on the overall saliency.
\item Salient \underline{k}ey\underline{w}ords (\textbf{\texttt{KW}}): given a passage, we ask LLMs to generate all salient keywords and hypernyms, as well as short phrases that may be inquired by users, focusing on precise saliency.
\end{itemize}

\noindent For passages, we prepare two different settings:

\begin{itemize}[noitemsep,nolistsep,leftmargin=*]
\item Out-of-domain (\textbf{OOD}): we sample 20,000 passages from existing resources such as DuReader, mostly consisting of web articles and titles.
\item In-domain (\textbf{ID}): we collect more image captions as the training passage pool. To mitigate memorization, we filter out all captions of ROUGE-L $> 0.6$ w.r.t. any test captions in CapRetrieval.
\end{itemize}

\begin{table}[tbp!]
\centering
\resizebox{0.95\columnwidth}{!}{
\begin{tabular}{ll|c|cc}
\toprule
&& \bf CapRetr & EcomRetr & VideoRetr \\
\midrule
& BGE & 78.86 & \bf 64.55 & \bf 69.91 \\
\midrule
\multirow{3}{*}{OOD} & \tt SM &  84.74 & 63.26 & 68.69 \\
& \tt KW & 87.23 & 60.49 & 63.82 \\
& \tt SM+KW & 86.46 & 60.91 & 64.89 \\
\midrule
\multirow{3}{*}{ID} & \tt SM & 84.61 & 62.45 & 67.47 \\
& \tt KW & 88.57 & 58.26 & 61.58 \\
& \tt SM+KW & \bf 91.83 & 60.24 & 65.16 \\
\bottomrule
\end{tabular}}
\caption{nDCG@10 on Cap/Ecom/Video-Retrieval. The same BGE encoder is continuously trained by using \texttt{SM} or \texttt{KW} or both as training queries, on the according OOD or ID corpus respectively (see Sec.~\ref{ssec:train-data} for acronyms).}
\label{tab:train-result}
\vspace{-1.5ex}
\end{table}

\noindent Table~\ref{tab:gen-example} illustrates training queries on an in-domain passage by the data generation strategies.

\begin{table*}[tbp!]
\centering
\resizebox{0.95\textwidth}{!}{
\begin{tabular}{c|c}
\multicolumn{2}{c}{\makecell[tc]{\textbf{Passage}: \cjk{图片显示了上海市电力公司的月度账单，包括2021年5月至9月的用电费用和支付状态。} \\ (\emph{The image shows the monthly bill from the Shanghai Electric Power Company,} \\ \emph{ including electricity charges and payment status from May to September 2021.})}} \\
\toprule
\tt \textbf{KW} (keywords/phrases) & \tt \textbf{SM} (summaries/queries) \\
\midrule
\makecell[tl]{Bill\\Electricity fee\\Utility bill\\Payment status\\ Monthly electricity charges\\Utility payment screenshot\\2021 electricity fee} &
\makecell[tl]{Shanghai Electric Power Company bill inquiry\\Electricity usage details from May to September 2021\\ Monthly electricity bill details\\Electricity payment status records\\ Historical payment information from Shanghai Electric Power Company\\Shanghai Electric Power Company electricity bill from May to September 2021\\...} \\
\end{tabular}}
\caption{Example of training queries on an in-domain passage, generated by our data generation strategies (Sec.~\ref{ssec:train-data}). \texttt{KW} strengthens the full view on precise keywords and concepts, while \texttt{SM} focuses on overall semantic saliency.}
\label{tab:gen-example}
\vspace{-1.5ex}
\end{table*}

\paragraph{Experimental Settings}
We finetune BGE 0.1B as the backbone encoder and use CLS token for embeddings. The training follows the typical InfoNCE contrastive loss \cite{infonce} with in-batch negatives. Training set statistics and more implementation details are provided in Appx.~\ref{app:train}.

For evaluation, we also include EcomRetrieval and VideoRetrieval \cite{multicpr} from MTEB, of which the passages are product/video titles, featuring similar lengths as image captions. As our experiments are conducted to examine the effect of training query granularity, we do not aim for other datasets nor general SOTA performance.

\subsection{The Granularity Dilemma}
\label{ssec:dilemma}

Table~\ref{tab:train-result} shows the results of the six training settings.\LN For CapRetrieval, the encoder trained with both \texttt{SM} and \texttt{KW} on in-domain passages achieves state-of-the-art performance, surpassing the original BGE significantly by 13\%, and outperforms the best baseline GTE-7B by 5+\%. For Ecom and Video, encoders trained with \texttt{SM} are shown comparable with BGE, indicating that summary-based queries share similar characteristics to existing large-scale training sets.
For CapRetrieval, models trained on in-domain corpus outperform those on OOD, while OOD models generalize better on Ecom and Video.

\textbf{Coarse vs. Fine}:
we roughly regard summaries and questions as coarse-grained queries that grasp important text semantics, resembling the existing training paradigm of most open-source encoders. Keywords/phrases on the other hand, are deemed fine-grained to strengthen the full semantic view of precise entities or concepts. However, Table~\ref{tab:train-result} suggests that while keywords drive substantial enhancement towards the entity and event retrieval, as shown by the clear improvement on CapRetrieval in both OOD and ID settings, they appear contributing little to Ecom and Video. 

Upon further analysis, we identified that though training with keywords could facilitate fine-grained matching, it may overlook the overall saliency. For example, the query in VideoRetrieval ``\cjk{荒野独居第2季中文版}'' (\emph{Alone Season 2 Chinese Edition}) should retrieve the TV show \emph{Alone} with the specified requirement; whereas the encoder trained with \texttt{KW} may overemphasize terms on \emph{Season 2} or \emph{Chinese}, but fails to correctly prioritize the supposedly most critical concept \emph{Alone}, showing a \textbf{misalignment of semantic importance}. 
We refer to this observation as the \emph{granularity dilemma}.

We hypothesize the main reason is that, the semantic importance of those precise keywords in a passage is relative, e.g. \emph{Season 2} is arguably less significant when accompanied with \emph{Alone}.
The current \texttt{KW} setting strengthens precise matching locally, but lacks training signals of fine-grained importance among keywords. This issue is not as severe for current open-source encoders, as their training queries comprise large-scale real-world queries that reflect user intents, which can implicitly curate importance through user clicking. However, for LLM-generated queries, it requires more engineering efforts to combat this issue.
We reckon that further analysis on training dynamics and data composition are needed to resolve this dilemma.

\section{Conclusion}

We focus on the embedding granularity that stems from the observation, where text encoders can fail to recognize entities or events of even simple cases.
A new evaluation set is introduced to probe the limitation of embeddings, and zero-shot experiments reveal the room for improvement on these fine-grained matching.
We further investigate data generation strategies for encoder training, and identify the \emph{granularity dilemma} that calls for future efforts towards more expressive embeddings.

\section*{Limitations}

As this work focuses around the granularity problem of embeddings, and examines both the zero-shot evaluation and training strategies, there can be limitations regarding the following two aspects.

First, the conducted analysis and training involve single-embedding encoders but exclude other paradigms, such as ColBERT \cite{colbert,colbertv2} or multi-views \cite{zhang-etal-2022-multi}. Addressing the dynamics beyond single embeddings can be important for an enlarged scope on this topic.

Second, it is still an open question on how to fully resolve the \emph{granularity dilemma}. As discussed in Sec.~\ref{ssec:dilemma}, we do provide our hypothesis on the keyword relative importance, and we leave the further investigation on training data composition outside the scope of this work.

\section*{Ethical Considerations}

For the dataset introduced in this work, we have manually reviewed each case to ensure compliance with privacy and ethical standards, in accordance with ACL ethics guidelines.
All passages and queries do not contain sensitive or biased content related to diversity or political viewpoints.
Personally identifiable information was anonymized, except in cases involving public figures where such information is part of the public domain.

No external annotators were recruited or employed during the dataset creation process; all annotation and verification were conducted internally by in-house researchers of this work. The data preparation and annotation process is approved and audited by the in-house research department.
All passages are derived from image transcriptions generated from GPT-4o. As such, there are no visible risks, copyright or legal concerns in using this dataset.

\bibliography{custom}

\clearpage
\appendix
\section{Related Work}
\label{app:related}

Dense retrieval serves a critical role and has received growing attention in recent developments of Retrieval-Augmented Generation (RAG) \cite{rag,raptor,xu-etal-2024-fine,graphrag,memorag,comorag}.
As a fundamental direction in representation learning, early works such as S-BERT \cite{sbert}, SimCSE \cite{simcse} and Contriever \cite{contriever} establish the effective training paradigm for embedding-based text representation, using contrastive learning on unsupervised or weakly supervised text pairs. Current state-of-the-art encoders usually adopt a multi-stage training that consists of both unsupervised and supervised finetuning stages \cite{openai-emb,e5,gte,bge}. Among the popular supervised training resources, most of the datasets are collected through real-world user behaviors, such as MSMARCO \cite{msmarco} and DuReader \cite{dureader-qa,dureader-rt}. Recently, synthetic data generation by LLMs is also reported positive gains in encoder training \cite{conan-v1,e5-mistral,yang2025qwen3technicalreport}.

Beyond the conventional single-embedding encoders, other paradigms have been proposed for retrieval, such as ColBERT with token-level embeddings \cite{colbert,colbertv2}, hybrid encoders with lexical features \cite{spar,hybrid-retriever,light-hybrid} and sparse features \cite{m3}. Recent efforts have also explored connecting global dependencies within embeddings \cite{sitemb}.

As far as our knowledge, we are the first to focus on the in-depth analysis on the embedding granularity problem, with a newly introduced dataset and controlled experiments on the 
encoder evaluation and training data strategies.

\section{Dataset}
\label{app:dataset}

The dataset is publicly released under the Apache 2.0 License.

\subsection{Query Types}
\label{app:query-type}

\begin{figure}[htp!]
\centering
\includegraphics[width=0.85\columnwidth]{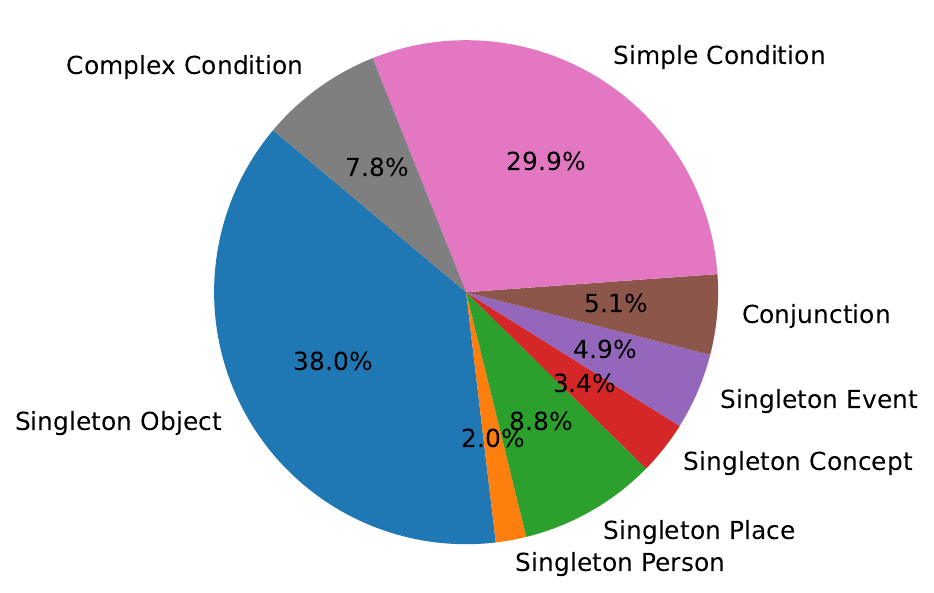}
\caption{Query types in CapRetrieval (details in \ref{app:query-type}).}
\label{fig:dataset-types}
\vspace{-1em}
\end{figure}

\begin{itemize}[noitemsep]
    \item \emph{Singleton Person}: person-related queries, e.g. \cjk{男性} (\emph{male}), \cjk{学生} (\emph{student}).
    \item \emph{Singleton Place}: place-related queries, e.g. \cjk{健身房} (\emph{gym}), \cjk{沙漠} (\emph{desert}).
    \item \emph{Singleton Object}: other concrete entities, e.g. \cjk{食物} (\emph{food}), \cjk{聊天记录} (\emph{chat history}).
    \item \emph{Singleton Concept}: non-concrete concepts, e.g. \cjk{音乐} (\emph{music}), \cjk{股票} (\emph{stocks}).
    \item \emph{Singleton Event}: event/action-related queries, e.g. \cjk{婚礼} (\emph{wedding}), \cjk{演唱会} (\emph{concert}).
    \item \emph{Conjunction}: conjuncted entities, e.g. \cjk{烧烤加啤酒} (\emph{BBQ and beer}), \cjk{樱花和传统建筑} (\emph{cherry blossoms and traditional building}).
    \item \emph{Simple Condition}: entities/events with simple conditions, e.g. \cjk{演唱会相关群聊} (\emph{group chat regarding concerts}), \cjk{睡觉的婴儿} (\emph{a sleeping baby}).
    \item \emph{Complex Condition}: entities/events with more complex conditions, e.g. \cjk{一个人在田地里收割白菜} (\emph{a person harvesting cabbages in the field}), \cjk{在沙发上的白猫} (\emph{a white cat next to a sofa}).
\end{itemize}

\noindent The distribution of query types is shown in Fig.~\ref{fig:dataset-types}.

\subsection{Label Distribution}

\begin{figure}[tp!]
\centering
\includegraphics[width=0.72\columnwidth]{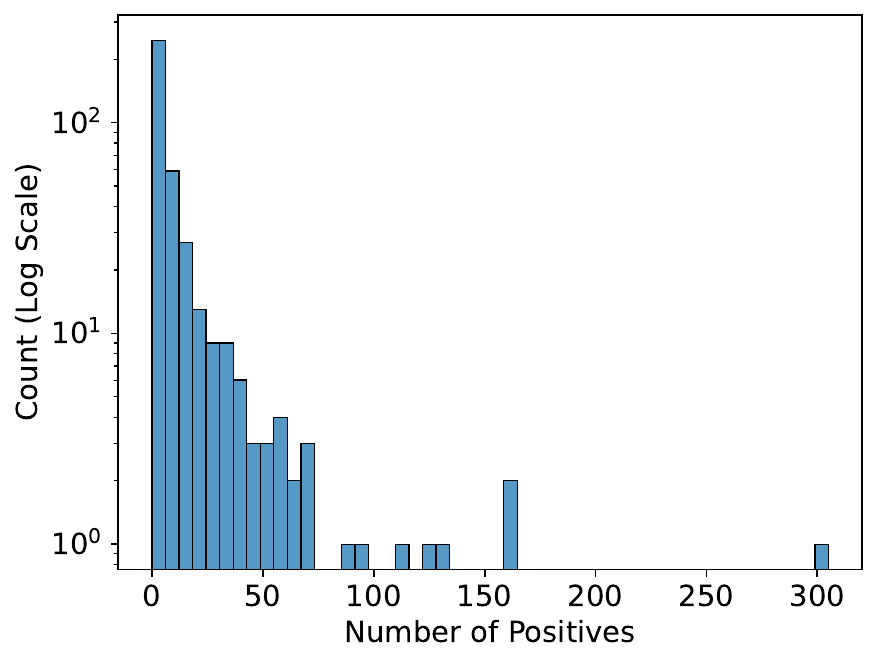}
\caption{Histogram of the number of relevant captions (labels of $[1,2]$) per query.}
\label{fig:dataset-pos}
\end{figure}

As mentioned in Sec.~\ref{sec:dataset}, full labels are annotated for each query-passage pair in CapRetrieval, resulting in a total number of 1.3 million pair labels, comprising 4,683 positive pairs. The distribution of positive passages per query is provided in Fig.~\ref{fig:dataset-pos}. Queries with the most number of relevant captions are general queries such as ``\cjk{男性}'' (\emph{male}), ``\cjk{女性}'' (\emph{female}), ``\cjk{食物}'' (\emph{food}).

Distinguished from typical retrieval datasets, we allow queries with no positive passages in CapRetrieval. There are 27 such queries out of the total 404 queries, and they are excluded for ranking-based metrics, i.e. nDCG scores in Table~\ref{tab:zero-result} does not consider these queries. However, they can be helpful examining encoder performance in a classification setting or adversarial analysis, which is supported by CapRetrieval, since all query-passage pairs are annotated.

\subsection{Annotation Procedure}

Two in-house researchers of this work participate in the annotation process; no external annotators are recruited. The whole procedure is conducted by a double annotation workflow.

\begin{enumerate}[noitemsep]
\item For each caption, each annotator independently assigns its labels for all the queries, such that each query-passage pair receives annotations from two annotators.
\item All label conflicts are collected and discussed by two annotators. The final label is either determined after reaching an agreement, or selected by the highest relevance received.
\item A round of correction is conducted by examining the retrieval results of preliminary experiments. Labels may be corrected after undergoing the same discussions by annotators.
\end{enumerate}

The first two steps of the annotation process took around 96 working hours in total, and the third step took an additional 30 hours. The final annotator agreement across relevant pairs is 95.7\%.

\section{Zero-Shot Evaluation}
\label{app:zero-shot}

For embedding-based retrieval, all experiments follows the same setting: all text is firstly converted to lower case; all embeddings are normalized such that cosine similarity is the metric for retrieval. For the GTE-1.5B/7B, E5-7B and Qwen3 models that take a query instruction, we use the following instruction ``\emph{Given an image search query, retrieve relevant image captions}'' in obtaining the query embedding for CapRetrieval.

Table~\ref{tab:zero-result-en} additionally provides zero-shot evaluation scores on the English version of CapRetrieval (dubbed CapRetrievalEn). GTE-7B achieves the best performance in both Chinese and English.

For BM25, we use \emph{Jieba}\footnote{\url{https://github.com/fxsjy/jieba}} for Chinese word segmentation and NLTK\footnote{\url{https://www.nltk.org}} for English tokenization. \emph{rank-bm25}\footnote{\url{https://pypi.org/project/rank-bm25}} is used for the Python BM25 implementation.

\begin{table}[tbp!]
\centering
\resizebox{\columnwidth}{!}{
\begin{tabular}{ll|ccc}
\toprule
&& nDCG@1 & nDCG@5 & \bf nDCG@10 \\
\midrule
& \it BM25 & 72.68 & 70.30 & 69.56 \\
\midrule
\multirow{3}{*}{0.1B} & BGE & 70.16 & 66.81 & 67.26 \\
& GTE & 78.12 & 76.60 & 75.77 \\
& E5 & 77.72 & 75.52 & 74.53 \\
\midrule
\multirow{2}{*}{0.3B} & BGE & 63.53 & 61.89 & 61.94 \\
& E5 & 80.77 & 77.90 & 77.40 \\
\midrule
\multirow{1}{*}{0.6B} & Qwen3 & 80.64 & 78.36 & 77.80 \\
\midrule
\multirow{4}{*}{> 1B} & GTE-1.5B & 73.21 & 72.46 & 72.04 \\
& GTE-7B & \bf 86.07 & \bf 83.70 & \bf 83.38 \\
& E5-7B & 78.91 & 77.25 & 77.07 \\
& Qwen3-8B & 80.11 & 78.83 & 78.38 \\
\bottomrule
\end{tabular}}
\caption{Evaluation results of zero-shot experiments on CapRetrievalEn (the according English version). The same multilingual encoder models are evaluated as in Table~\ref{tab:zero-result}, except for BGE that is now switched to its dedicated English version.}
\label{tab:zero-result-en}
\end{table}

\section{Encoder Training}
\label{app:train}

\begin{table}[htbp!]
\centering
\resizebox{0.95\columnwidth}{!}{
\begin{tabular}{ll|cc|cc}
\toprule
&& Passages & Tokens & Queries & Length \\
\midrule
\multirow{2}{*}{OOD} & \tt SM & 20K & 2.8M & 7.0 & 13.7 \\
& \tt KW & 20K & 2.8M & 13.9 & 6.0 \\
\midrule
\multirow{2}{*}{ID} & \tt SM & 40K & 1.5M & 7.0 & 11.6 \\
& \tt KW & 40K & 1.5M & 12.2 & 4.0 \\
\bottomrule
\end{tabular}}
\caption{Statistics of the training set settings described in Sec.~\ref{ssec:train-data}: number of passages, total tokens of passages, averaged number of generated queries per passage, averaged number of tokens per query.}
\label{tab:train-stats}
\end{table}

Table~\ref{tab:train-stats} shows the statistics of the training set settings described in Sec.~\ref{ssec:train-data}, and 5\% queries are randomly sampled as the holdout set. We continuously train \emph{bge-base-zh-v1.5} on a single Nvidia GPU with the typical InfoNCE contrastive loss, learning rate $5 \times 10^{-6}$, weight decay $0.1$, temperature $0.01$. The number of epochs and batch size is adjusted to derive around 4K training steps.

The evaluation results in Table~\ref{tab:train-result} demonstrate the effectiveness of our proposed data generation strategies on strengthening the fine-grained embedding matching, surpassing the baseline using either in-domain corpus or out-of-domain passages. Through cross-examination, we further raise the \emph{granularity dilemma} discussed in Sec.~\ref{ssec:dilemma}.

\begin{table*}[thbp!]
\centering
\resizebox{\textwidth}{!}{
\begin{tabular}{c|c|c|c}
\toprule
Query & Passages & Label & Similarity \\
\midrule
\multirow{6}{*}{\makecell{\cjk{炸鸡}\\(\emph{fried}\\ \emph{chicken})}} & \makecell[tl]{\textcolor{purple}{\cjk{一桌丰盛的餐点包括烤肉串、炸薯条和春卷。}} \\ (\emph{A table full of delicious dishes includes grilled meat skewers,}\\ \emph{ French fries, and spring rolls.})} & \multirow{3}{*}{\textcolor{purple}{0}} & \multirow{3}{*}{0.48} \\
&&\\
& \makecell[tl]{\textcolor{teal}{\cjk{图片展示了麦当劳麦辣鸡翅（2块）20次券的电子优惠券，售价185.3元，}}\\ \textcolor{teal}{\cjk{单份低至7.9元。}} \\ (\emph{The image shows a digital coupon for 20 servings of McDonald's Spicy Chicken Wings}\\\emph{ (2 pieces), priced at 185.3 RMB, bringing the cost as low as 7.9 RMB per serving.})} & \multirow{4}{*}{\textcolor{teal}{2}} & \multirow{4}{*}{0.38} \\
\midrule
\multirow{14}{*}{\makecell{\cjk{紫色的花}\\(\emph{purple}\\ \emph{flower})}} & \makecell[tl]{\textcolor{purple}{\cjk{一辆紫色轿车停在路边，车顶和车窗上装饰有花束，车前挡风玻璃上有红色标签。}} \\ (\emph{A purple sedan is parked by the roadside, decorated with flower bouquets on the roof}\\ \emph{ and windows, and a red tag on the front windshield.})} & \multirow{3}{*}{\textcolor{purple}{0}} & \multirow{3}{*}{0.57} \\
&&\\
& \makecell[tl]{\textcolor{purple}{\cjk{图片中有四只紫色的蝴蝶，背景为浅紫色。}} \\ (\emph{The image features four purple butterflies against a light purple background.})} & \multirow{2}{*}{\textcolor{purple}{0}} & \multirow{2}{*}{0.57} \\
&&\\
& \makecell[tl]{\textcolor{teal}{\cjk{一辆白色轿车停在树下，背景是紫色花田和远处的山脉。}} \\ (\emph{A white sedan is parked under a tree, with a purple flower field and}\\ \emph{ distant mountains in the background.})} & \multirow{3}{*}{\textcolor{teal}{2}} & \multirow{3}{*}{0.48} \\
&&\\
& \makecell[tl]{\textcolor{teal}{\cjk{图片展示了一片薰衣草田，背景是蓝天白云，文字内容为“只要你欢乐，}}\\ \textcolor{teal}{\cjk{我就幸福浓浓。朋友，愿你欢乐无忧。早上好”。}} \\ (\emph{The image shows a lavender field with a background of blue sky and white clouds.}\\ \emph{ The text reads: ``As long as you're happy, my heart is full of joy. My friend,}\\ \emph{ may you be cheerful and carefree. Good morning.''})} & \multirow{5}{*}{\textcolor{teal}{2}} & \multirow{5}{*}{0.37} \\
\bottomrule
\end{tabular}}
\caption{Examples on dense retrieval selected from the zero-shot experiments on our new evaluation set CapRetrieval. Passages in \textcolor{purple}{Red} are labeled \textcolor{purple}{irrelevant} to queries (label 0), and passages in \textcolor{teal}{Green} are \textcolor{teal}{relevant} (label 2). For both queries, encoders retrieve irrelevant passages before the more relevant ones, despite all queries and passages are straightforward to comprehend. The cosine similarity is provided rightmost using the popular open-source encoder for Chinese \emph{bge-large-zh-v1.5}. However, it should be noted that all encoders from both the popular BGE and GTE encoder series fail on the above examples, spanning from 0.1B to even 7B models. Overall, encoders can exhibit flaws on fine-grained embedding matching even on simple cases, regardless of training sources and model sizes.}
\label{tab:example-a}
\end{table*}

\begin{table*}[thbp!]
\centering
\resizebox{0.9\textwidth}{!}{
\begin{tabular}{c|c|c|c}
\toprule
Query & Passages & Label & Similarity \\
\midrule
\multirow{3}{*}{{\makecell{\cjk{西瓜}\\(\emph{watermelon})}}} & \makecell[tl]{\textcolor{purple}{\cjk{图片中有一个装满水果的篮子，旁边有生菜、猕猴桃和小番茄。}}} & \textcolor{purple}{0} & 0.50 \\
&&\\
& \makecell[tl]{\textcolor{teal}{\cjk{一辆装满西瓜的三轮车停在商店门口。}}} & \textcolor{teal}{2} & 0.47 \\
\midrule
\multirow{4}{*}{watermelon} & \makecell[tl]{\textcolor{purple}{In the picture, there is a basket full of fruit, with lettuce, kiwis, and}\\ \textcolor{purple}{cherry tomatoes next to it.}} & \multirow{2}{*}{\textcolor{purple}{0}} & \multirow{2}{*}{0.60} \\
&&\\
& \makecell[tl]{\textcolor{teal}{A tricycle loaded with watermelons is parked in front of the store.}} & \textcolor{teal}{2} & 0.55 \\
\bottomrule
\end{tabular}}
\caption{A more extreme example to illustrate the embedding granularity problem. For the query \cjk{西瓜} (\emph{watermelon}) and passages in both Chinese and English accordingly, both the popular BGE large encoders \emph{bge-large-zh/en-v1.5} fail to retrieve the obviously more \textcolor{teal}{relevant} passage (label 2) before the \textcolor{purple}{irrelevant} one (label 0). Though this case can be simply resolved by lexical matching, it demonstrates that the embedding granularity problem exists across languages. In this work, we primarily focus on the regarding evaluation in Chinese.}
\label{tab:example-b}
\end{table*}

\end{document}